\documentclass[11pt]{article}
\pdfoutput=1

\usepackage{hyperref}

\makeatletter

\def\showhyphens#1{}%
\makeatother

\usepackage[final]{acl}

\usepackage{times}
\usepackage{latexsym}
\usepackage{graphicx}
\usepackage{placeins}
\usepackage{natbib}
\usepackage{stfloats} 
\usepackage{lipsum}
\usepackage{booktabs}
\usepackage{amssymb}
\usepackage[T1]{fontenc}
\usepackage[utf8]{inputenc}
\usepackage{microtype}

\usepackage{inconsolata}

\usepackage{amsmath}
\usepackage{accents}

\title{ReasonScaffold: A Scaffolded Reasoning-based Annotation Protocol for Human-AI Co-Annotation}

\author{Smitha Muthya Sudheendra\\
  University of Minnesota, Twin Cities\\
  \texttt{muthy009@umn.edu} \\\And
  \And
  \texttt{Jaideep Srivastava} \\
  University of Minnesota, Twin Cities\\
  \texttt{srivasta@umn.edu}}

\begin{document}
\maketitle


\begin{abstract}
Human annotation is central to NLP evaluation, yet subjective tasks often exhibit substantial variability across annotators. While large language models (LLMs) can provide structured reasoning to support annotation, their influence on human annotation behavior remains underexplored. We introduce \textbf{ReasonScaffold}, a scaffolded reasoning annotation protocol that exposes LLM-generated explanations while withholding predicted labels. We study how reasoning affects human annotation behavior in a controlled setting, rather than evaluating annotation accuracy. Using a two-pass protocol inspired by Delphi-style revision, annotators first label instances independently and then revise their decisions after viewing model-generated reasoning. We evaluate the approach on sentiment classification and opinion detection tasks, analyzing changes in inter-annotator agreement and revision behavior. To quantify these effects, we introduce the Annotator Effort Proxy (AEP), a metric capturing the proportion of labels revised after exposure to reasoning. Our results show that exposure to reasoning is associated with increased agreement, along with minimal revision, suggesting that reasoning helps resolve ambiguous cases without inducing widespread changes. These findings provide insight into how reasoning explanations shape annotation consistency and highlight reasoning-based scaffolds as a practical mechanism for human--AI co-annotation workflows.
\end{abstract}

\section{Introduction}
Reliable evaluation practices are central to building effective natural language processing (NLP) systems, especially in settings that rely on human annotation. While high quality labeled data is essential for training and evaluating models, annotating subjective language often leads to substantial variability across annotators. Tasks such as sentiment analysis and opinion detection involve subtle cues and context dependent interpretations, so even well-defined guidelines do not always prevent disagreement. From an evaluation standpoint, this variability is not just noise, it reflects the inherent ambiguity of the task and complicates how we assess both data quality and downstream model performance.

Recent advances in large language models (LLMs) offer new ways to support annotation workflows. Beyond generating labels, LLMs can produce structured reasoning that highlights relevant linguistic cues and plausible interpretations of an utterance. In principle, these explanations can help annotators better understand ambiguous cases. In practice, however, many LLM-assisted annotation approaches rely on predicted labels or suggestions, which can unintentionally influence annotators through overreliance or anchoring effects.

This raises an important sociotechnical question: how can we design annotation workflows that benefit from model assistance without undermining human judgment? Understanding how model-generated reasoning affects annotator decisions is particularly important for dataset creation, where human interpretation ultimately defines the ground truth. To explore this question, we introduce \emph{ReasonScaffold}, a scaffolded reasoning-based annotation protocol that exposes LLM-generated explanations while withholding predicted labels. The system produces task-aligned reasoning via self-example prompting and chain-of-thought analysis, highlighting cues relevant to the annotation task. Rather than telling annotators what label to choose, these explanations act as interpretive scaffolds that annotators can use, question, or ignore, while retaining full control over their final decisions.

\begin{figure}[h]
  \centering
  \includegraphics[width=\linewidth]{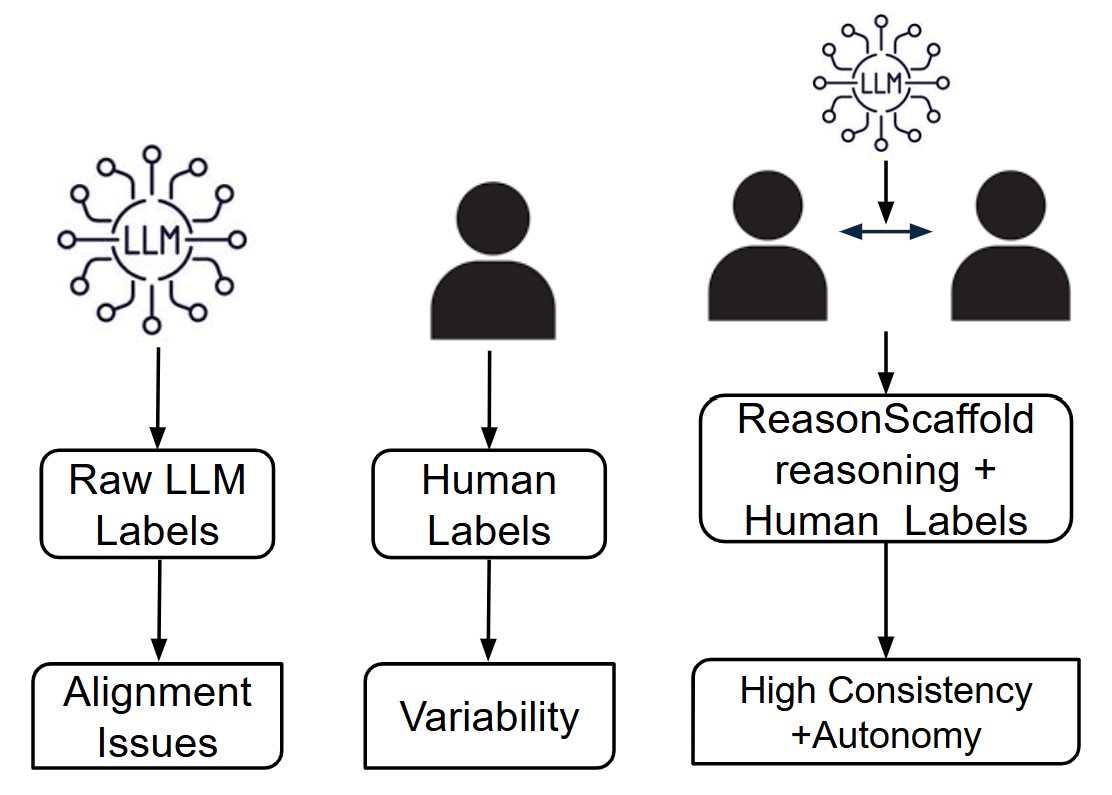}
  \caption{Comparison of Annotation Approaches: automated labeling, human-only annotation, and the ReasonScaffold}
  \label{annotation approaches}
\end{figure}

Figure~\ref{annotation approaches} illustrates how ReasonScaffold sits between two common annotation paradigms. Fully automated labeling with LLM predictions is fast but can misalign with human interpretation, while human annotation preserves contextual judgment but often results in variability. ReasonScaffold introduces a middle ground: a scaffolded reasoning-based annotation protocol that supports more consistent interpretation without overriding human autonomy. We use the term ReasonScaffold to emphasize that the protocol provides structured reasoning support for annotators rather than enforcing strict alignment with model outputs.

We study the effect of this scaffold using a controlled two-pass annotation protocol. Annotators first label each instance independently and then revisit their decisions after viewing the LLM-generated reasoning. This setup allows us to isolate the effect of exposure to reasoning on annotation behavior while limiting direct reliance on model outputs. In addition to measuring inter-annotator agreement, we analyze revision patterns to better understand how annotators respond to the reasoning scaffold.

Overall, this work contributes to evaluation methodology for Human--AI annotation workflows by examining how reasoning explanations affect agreement, revision behavior, and interpretive stability. Specifically, we make the following contributions:

\begin{itemize}
    \item Develop an experimental protocol to compare human annotations before and after exposure to LLM-generated reasoning.
    \item Quantify how LLM outputs influence human labeling decisions when discrepancies arise.
    \item Formalize the Annotator Effort Proxy (AEP) as a metric of collective annotator response to model explanations.
    \item Validate the protocol and metric on affective state detection, particularly sentiment analysis.
    \item Assess the risk of anchoring bias when LLM reasoning is introduced into annotation workflows.
\end{itemize}

\section{Background}
Human--AI collaborative annotation has gained traction as large language models (LLMs) become increasingly capable of supporting complex labeling tasks. Prior work has explored hybrid workflows that combine human judgment with model assistance to improve efficiency and scalability. For example, CoAnnotating \cite{li2023coannotating} proposes an uncertainty-guided framework that dynamically allocates instances between human annotators and models. Related approaches extend Human--AI co-annotation to specific domains, including collaborative annotation systems for online incivility
\cite{park2024collaborative} and LLM-assisted annotation frameworks for speech datasets \cite{johnson2025exploratory}, as well as human-centered annotation systems that balance automation with human oversight \cite{pangakis2024humansintheloop}.

While these methods improve efficiency and task allocation, they typically rely on model predictions or uncertainty signals to guide annotation. Consequently, they provide limited insight into how model-generated reasoning influences human interpretation, particularly in subjective tasks where multiple interpretations may be plausible.

Recent studies have highlighted both the potential benefits and risks of LLM-assisted annotation. Annotators have been shown to adopt model suggestions even when they are incorrect \cite{schroeder2025just}, and similar susceptibility to bias induced by AI-generated suggestions has been observed \cite{beck2025biasintheloop}. These findings indicate systematic shifts in annotator judgments and increased reliance on model outputs, which are especially problematic in subjective tasks with high interpretive variability.

More broadly, research on how humans interact with AI explanations shows that their effects are not straightforward: explanations can both help and hinder decision-making, depending on how they are presented. Prior work finds that explanations do not consistently improve complementary Human--AI performance \cite{bansal2021does} and may not reliably support appropriate reliance on model outputs \cite{fok2024search}. Recent work in NLP further shows that the usefulness of explanations in Human--AI decision-making is often inconsistent and highly sensitive to evaluation design and experimental context \cite{chaleshtori2024evaluating}. Together, these findings highlight the need for evaluation protocols that can isolate the specific influence of reasoning explanations from that of model predictions.

The development of chain-of-thought prompting \cite{wei2022chain} has shown that LLMs can generate structured, step-by-step reasoning that improves reasoning performance and produces intermediate explanations. Subsequent work suggests that explicit reasoning can better capture patterns of human subjectivity and variation in interpretation \cite{giorgi2024modeling}. However, most applications of chain-of-thought focus on improving model predictions rather than supporting human decision processes.

In this work, we repurpose chain-of-thought reasoning as an \emph{interpretive scaffold} for human annotators. By exposing reasoning without predicted labels, the scaffold provides interpretable cues without prescribing a final decision, allowing annotators to engage critically rather than adopt model outputs.

Our focus on sentiment and opinion classification builds on prior work on subjective language, which highlights how evaluative expressions are often subtle and highly dependent on context \cite{wiebe2004subjective, wilson2005contextual, mohammad2016sentiment}. As these judgments rely on interpretation, annotators may reasonably disagree, leading to variability in labels. This makes such tasks a useful setting for exploring methods that can improve consistency while still preserving diverse perspectives.

Our two-pass annotation protocol is conceptually related to structured expert revision methods such as the Delphi technique \cite{linstone1975delphi}, in which participants provide independent judgments and subsequently revise them after receiving structured feedback. The Delphi method has been widely used to promote consensus and reduce
variability in expert decision-making. In contrast to traditional Delphi settings, where feedback is derived from other human participants, our approach introduces LLM-generated reasoning as a structured interpretive signal. This can be viewed as an AI-augmented variant of Delphi-style annotation, where model-generated explanations replace peer feedback while preserving independence from model predictions.

Across these lines of work, prior approaches either delegate labeling decisions to models, provide suggestion-based assistance, or optimize task allocation between humans and AI systems. In contrast, our work focuses on evaluation methodology: we introduce a controlled two-pass protocol to study how reasoning explanations influence annotator agreement, revision behavior, and interpretive stability. By isolating reasoning from
prediction, we aim to better understand the role of LLM-generated explanations in Human--AI annotation workflows.

\section{ReasonScaffold Framework}
ReasonScaffold is a Human--AI co-annotation protocol that uses structured LLM-generated reasoning as an interpretive scaffold to support consistent human annotation while preserving annotator autonomy. The framework is motivated by the observation that annotation disagreements in subjective tasks often arise not from lack of task knowledge, but from divergent interpretations of subtle linguistic cues, implicit affective signals, or
ambiguous conversational context. Rather than replacing human judgment, ReasonScaffold makes these cues explicit, enabling more consistent reasoning across annotators while retaining full control over final labels.

\textbf{Phase 1: Independent Human Annotation:} The protocol begins with an unassisted first pass in which annotators label all instances independently using only task guidelines. This phase captures baseline interpretive variability and establishes a reference point for evaluating the effect of model-generated reasoning. Annotators are not informed of a subsequent revision phase to avoid anticipation effects or
strategic deferral of difficult cases.

\textbf{Phase 2: LLM-Based Interpretive Scaffold:} After the initial pass, the LLM generates an interpretive scaffold for each instance, a structured explanation of how the utterance may be interpreted with respect to task definitions. The scaffold is constructed using two complementary components. First, self-example prompting generates representative examples for each label category. These examples expose how the model internally defines category boundaries and provide concrete reference points for interpreting ambiguous cases. Because the examples are generated within the same prompt context, they maintain consistency with the model’s subsequent reasoning. Second, the model produces chain-of-thought reasoning (CoT), offering a step-by-step analysis of the utterance. This reasoning highlights lexical cues (e.g., hedging, intensifiers), pragmatic signals (e.g.,
indirect complaints), contextual dependencies, and distinctions between adjacent classes. Together, these components transform implicit model assumptions into explicit, inspectable reasoning that annotators can
engage with critically.

\textbf{Phase 3: Human Reflection with Reasoning:} In the final phase, annotators revisit their initial labels while viewing only the model’s reasoning. Importantly, predicted labels are not shown. Annotators are instructed to revise a label only if the explanation reveals previously overlooked cues or clarifies an ambiguous boundary. This design encourages interpretive reflection rather than answer adoption, allowing us to isolate the effect of reasoning exposure on annotation behavior. 

Revisions are observed across annotators and instances, suggesting that the scaffold supports interpretive recalibration rather than simple convergence toward model outputs. Changes are typically associated with newly surfaced implicit signals, contextual dependencies, or clearer distinctions between label categories.

\begin{figure}[h]
  \centering
  \includegraphics[width=\linewidth]{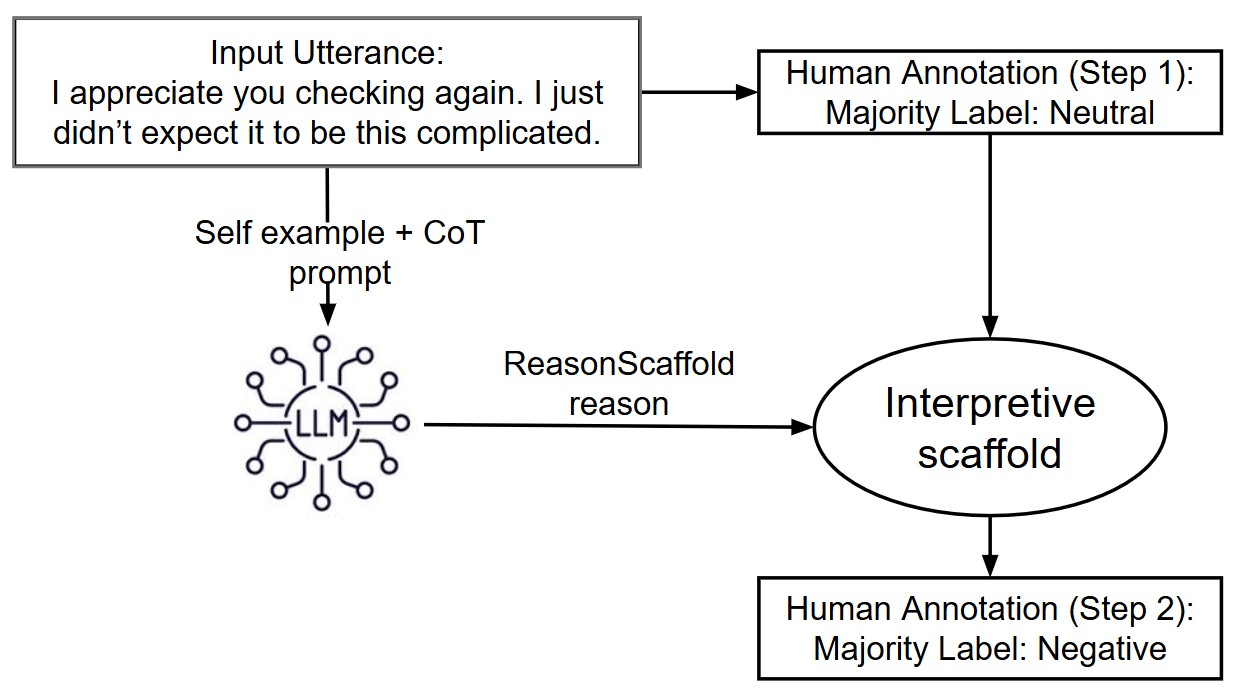}
  \caption{ReasonScaffold: Interpretive scaffold for Human--AI co-annotation.}
  \label{ISF}
\end{figure}

\textbf{Interpretive Loop:} Together, these phases form a controlled interpretive loop: independent human
judgment establishes a baseline, model-generated reasoning provides structured analytic support, and human revision produces the final labels. This design enables systematic analysis of how reasoning affects agreement
and revision behavior while preserving full human authorship.

\textbf{Safeguards Against Model-Induced Bias:} To mitigate model-induced bias, ReasonScaffold incorporates several safeguards. First, all annotations are completed independently before exposure to model reasoning. Second, annotators are shown only explanations and never model predictions, reducing anchoring effects common in suggestion-based systems. Finally, explanations are presented as fallible arguments rather than
authoritative outputs, and annotators are instructed to revise labels only when the reasoning provides meaningful new insight. These design choices ensure that observed changes reflect interpretive clarification rather than assimilation to model outputs.

\section{Experimental Setup}
We evaluate ReasonScaffold on conversational utterances to examine how structured LLM reasoning influences interpretive consistency in subjective annotation tasks. Conversational data provides a realistic test setting because it often contains spontaneous, mixed-intent utterances in which evaluative cues are subtle, indirect, or embedded within otherwise task-oriented statements. This makes it well-suited for studying how interpretive scaffolding affects human judgment. We consider two downstream tasks:
\begin{itemize}
    \item Sentiment Analysis: Identifying whether an utterance expresses a positive, negative, or neutral affective stance.
    \item Opinion Detection: Determining whether an utterance conveys a personal evaluation, judgment, preference, or stance.
\end{itemize}

These tasks were selected for three reasons. First, both are conceptually clear and grounded in established literature, which supports relatively consistent interpretation across annotators and the model. Second, both are representative subjective classification tasks in which disagreement often arises from subtle differences in interpretation rather than from lack of task understanding. Third, their relative simplicity makes them suitable for a controlled study of human--LLM interaction without confounds introduced by highly specialized domains or expert-only knowledge.

To generate model reasoning, we use the self-example prompting and chain-of-thought procedure described in Appendix A. Self-generated examples provide reference points for category boundaries, while chain-of-thought explanations articulate the model’s interpretive process for each instance. To assess both baseline human agreement and the influence of structured
reasoning, we implement a two-pass annotation protocol aligned with the ReasonScaffold design.

\emph{Pass 1: Human-Only Annotation:} Annotators label all utterances independently without exposure to any LLM output. This pass captures unassisted judgments based only on the task
guidelines and provides a baseline measure of interpretive variability. 

\emph{Pass 2: Human Annotation with Scaffolded Reasoning:} Annotators then review the model’s reasoning explanation for each instance and are asked to reconsider their initial labels. They are instructed to revise a label only when the explanation reveals relevant cues they had not previously considered or clarifies an ambiguous boundary. Importantly, annotators do not see the model’s predicted label.

This design reduces direct adoption of model outputs and allows us to focus specifically on the influence of reasoning explanations. This setup provides a controlled basis for analyzing whether exposure to structured reasoning is associated with changes in agreement, revision behavior, and interpretive stability. Although it does not eliminate all possible forms of influence, it isolates reasoning from prediction and therefore enables a more targeted evaluation of how model explanations shape human annotation decisions.

\section{Evaluation}
We evaluate ReasonScaffold to examine how exposure to LLM-generated reasoning affects annotation reliability, interpretive stability, and revision behavior in subjective annotation tasks. Our evaluation is designed as a controlled study to isolate the effect of reasoning explanations while preserving human autonomy.

Experiments are conducted on two widely used subjective classification tasks: sentiment classification and opinion detection over conversational utterances. These tasks involve context-dependent and often implicit cues, making them prone to annotator disagreement and therefore suitable for studying how structured reasoning influences interpretation.

\subsection{Dataset and Annotators}
A subset of 500 utterances was annotated by four annotators per task. Annotators were selected based on (i) prior annotation experience, (ii) familiarity with conversational language, and (iii) fluency in English. All annotators were provided with identical task definitions and examples. No additional calibration or discussion was conducted prior to annotation, allowing us to capture natural variability in interpretation. 

While limiting analysis of group-level variability, this setup enables a controlled examination of how exposure to LLM-generated reasoning affects pairwise annotation consistency. Although four annotators participated, agreement is computed pairwise to isolate interaction effects. This design prioritizes interpretability over large-scale variability.

\subsection{Experimental Protocol}
We adopt a two-pass annotation protocol aligned with the ReasonScaffold framework. 

\textbf{Pass 1: Independent Annotation:} Annotators label all utterances independently without access to any model output. This pass establishes a baseline measure of inter-annotator agreement and captures unassisted interpretive variability.

\textbf{Pass 2: Annotation with Reasoning Scaffold:} Annotators then revisit their initial labels after viewing the LLM-generated reasoning scaffold. The scaffold includes self-generated examples and structured reasoning but does not reveal model predictions. Annotators are instructed to revise labels only when the explanation introduces relevant cues or clarifies an ambiguous boundary. This design isolates the effect of reasoning explanations from direct label suggestion, enabling a more targeted evaluation of how interpretive support influences annotation behavior.

\subsection{Inter-Annotator Agreement}
We measure agreement using \textbf{Cohen's $\kappa$} across annotators. In the first pass, agreement averaged $\kappa = 0.76$ for sentiment and $\kappa = 0.73$ for opinion detection, indicating substantial but imperfect alignment.

After exposure to the reasoning scaffold, agreement increased to $\kappa = 0.98$ for sentiment and $\kappa = 0.85$ for opinion detection. These improvements suggest that structured reasoning helps reduce interpretive divergence, particularly in cases involving indirect sentiment, mixed polarity, or implicit evaluative cues.

While the observed gains are substantial, the near-perfect agreement in the sentiment task should be interpreted cautiously, given the limited dataset size and controlled experimental setting. Nevertheless, the consistent increase across both tasks indicates that reasoning exposure is associated with greater alignment in annotator decisions. However, the increased agreement does not necessarily imply improved correctness. In this study, we focus on interpretive alignment rather than ground-truth accuracy, which remains an important direction for future work.

\subsection{Annotator Effort Proxy (AEP)}
To quantify how exposure to LLM-generated reasoning affects annotation decisions, we introduce the \emph{Annotator Effort Proxy (AEP)}, defined as:

\begin{equation}
\mathrm{AEP} = \frac{\text{Number of revised labels}}{\text{Total labels}}
\end{equation}

AEP measures the proportion of labels modified between the first (independent) and second (reasoning-assisted) annotation passes. In a two-annotator setting, this metric complements Cohen’s $\kappa$ by capturing \emph{within-annotator} changes rather than agreement between annotators. As such, AEP provides a direct view of how reasoning exposure influences individual annotation behavior. Low AEP values indicate that annotators largely retain their initial judgments, suggesting that reasoning reinforces or clarifies existing interpretations rather than overriding them. Higher AEP values, in contrast, would indicate substantial shifts in labeling decisions.

In our experiments, AEP values are low: 0.74\% for sentiment classification and 1.05\% for opinion detection. Revisions occur in both directions (e.g., negative to positive, and vice versa; opinion to non-opinion and vice versa), indicating selective reconsideration rather than systematic convergence toward a particular outcome.

When considered alongside increases in Cohen’s $\kappa$, these results suggest that reasoning improves pairwise annotation consistency primarily by clarifying ambiguous cases rather than inducing widespread label changes. In other words, annotators converge more often not because they change many labels, but because they resolve a small number of difficult cases in a consistent way.

Importantly, AEP should not be interpreted as a measure of annotation effort or time. In this work, it serves specifically as an indicator of interpretive stability and the extent to which reasoning prompts meaningful revision.

\begin{table}[h]
\centering
\small
\begin{tabular}{lcccc}
\hline
\textbf{Task} & \textbf{$\kappa_1$} & \textbf{$\kappa_2$} & \textbf{AEP (\%)}\\
\hline
Sentiment & 0.76 & 0.98 & 0.74 \\
Opinion   & 0.73 & 0.85 & 1.05 \\
\hline
\end{tabular}
\caption{Summary of annotation behavior across passes.}
\label{tab:full_eval}
\end{table}

Table~\ref{tab:full_eval} summarizes agreement and revision behavior across both annotation passes. The combination of increased $\kappa$ and low AEP suggests that improvements in agreement are driven by resolution of a small number of ambiguous cases rather than widespread label changes. $\kappa_1$ and $\kappa_2$ denote Cohen’s $\kappa$ before and after exposure to reasoning and AEP measures revision rate.
\subsection{Disagreement Resolution Analysis}

To further understand how reasoning affects annotation behavior, we analyze instances where the two annotators disagreed in the first pass and examine whether these disagreements are resolved after exposure to the reasoning scaffold. We track:
\begin{itemize}
    \item the number of items with disagreement in the first pass,
    \item the number of those items that reach agreement in the second pass,
    \item and the number that remain unresolved.
\end{itemize}
A substantial portion of initially disputed items reach agreement after annotators review the reasoning scaffold, suggesting that the explanations help surface relevant cues and reduce ambiguity. While we do not quantify this proportion explicitly, the observed trend is consistent across both tasks. These cases often involve implicit sentiment, indirect evaluation, or context-dependent interpretations that are not immediately apparent in the first pass. Remaining disagreements typically correspond to inherently ambiguous instances where multiple interpretations remain plausible even after reasoning is provided.

\subsection{Anchoring and Influence Considerations}
A key concern in LLM-assisted annotation is whether exposure to model outputs induces anchoring or overreliance. In our setup, annotators are exposed only to model-generated reasoning and not predicted labels, reducing the strongest source of anchoring observed in suggestion-based workflows.

Empirically, revision rates are low ($<1.1\%$) and bidirectional, which is inconsistent with the unidirectional drift typically associated with anchoring. In addition, Cohen’s $\kappa$ measures agreement between annotators rather than similarity to model outputs, so increases in agreement cannot be directly attributed to model dominance.

Taken together, these observations suggest that annotators engage with the reasoning scaffold selectively and critically. However, we do not directly measure anchoring effects, and future work should examine this more explicitly through controlled interventions.
\subsection{Consistency and Reliability of LLM Reasoning}
While annotators are not exposed to model predictions in the ReasonScaffold protocol, we analyze the model’s underlying probabilistic outputs offline to assess the stability and reliability of its interpretive behavior. This allows us to verify that the reasoning scaffold is supported by consistent model signals rather than stochastic variation.

To evaluate consistency, we re-prompt the model on a subset of 100 utterances across three independent runs under identical conditions (temperature = 0.2). The resulting soft label probabilities exhibit a high mean inter-run correlation of $r = 0.94$, indicating strong internal stability in the model’s responses.

We further assess alignment between model signals and human annotations using the Brier score, obtaining values of 0.09 for sentiment and 0.12 for opinion detection. These results suggest that the model’s underlying probabilistic estimates are well calibrated with respect to human consensus.

Importantly, these analyses are conducted offline and do not influence the annotation process, which relies solely on reasoning explanations. Together, these findings support the use of LLM-generated reasoning as a stable and reproducible interpretive aid in annotation workflows.

\subsection{Reasoning Quality Considerations}
This study focuses on how exposure to structured LLM-generated reasoning affects human annotation behavior rather than evaluating the intrinsic correctness of the reasoning itself. Reasoning quality is therefore assessed indirectly through annotator responses.

The low revision rates and absence of systematic directional shifts suggest that annotators do not simply adopt model interpretations. Instead, they appear to use the reasoning as an interpretive aid, revising labels only when new or previously overlooked cues are identified. A more direct evaluation of reasoning correctness and its relationship to annotation outcomes remains an important direction for future work.
\section{Application of ReasonScaffold}
From a practical standpoint, ReasonScaffold can be integrated into annotation pipelines with minimal overhead, as it relies on post-hoc reasoning generation rather than real-time model predictions. Given the low revision rates observed (<1.1\%), annotators are not required to substantially rework labels, suggesting that the protocol can be incorporated into existing workflows with limited additional cost.

In practice, this approach is particularly well-suited for high-ambiguity tasks, where small improvements in agreement can meaningfully enhance dataset consistency. While we do not explicitly measure annotation time, the low revision rates indicate that the second pass primarily affects a small subset of ambiguous instances, suggesting a modest additional time cost relative to potential gains in agreement. We envision ReasonScaffold being most useful in early-stage dataset construction and in subjective labeling tasks where interpretive ambiguity is a primary source of disagreement.

\section{Conclusion}
In this work, we introduced ReasonScaffold, a scaffolded reasoning-based annotation protocol designed to examine how LLM-generated explanations influence human annotation behavior. Through a controlled two-pass evaluation, we studied the role of structured reasoning in shaping annotation reliability, interpretive stability, and revision patterns in subjective classification tasks.

Our findings show that exposure to model-generated reasoning is associated with increased inter-annotator agreement and selective revision of labels, particularly in cases involving implicit or ambiguous cues. Importantly, these improvements occur alongside low revision rates and bidirectional label changes, suggesting that reasoning functions as an interpretive aid rather than a prescriptive signal. This supports the view that structured explanations can help clarify decision boundaries without undermining annotator autonomy.

Beyond empirical results, this work contributes a practical evaluation framework for studying human–AI interaction in annotation workflows. By isolating reasoning from prediction and introducing complementary metrics such as the Annotator Effort Proxy (AEP), our approach provides a more granular understanding of how model outputs influence human decision-making. This perspective is particularly relevant for dataset construction, where annotation processes directly shape downstream model behavior.

Overall, ReasonScaffold highlights the potential of reasoning-based scaffolds as a principled mechanism for improving annotation consistency while preserving human interpretive control.

\section{Limitations}

This study has several limitations that point to directions for future work. The evaluation is conducted on a relatively small dataset with a limited number of annotators, and reasoning quality is assessed indirectly through annotator responses rather than explicit correctness measures. Additionally, while the design reduces direct anchoring effects, it does not fully eliminate all forms of model influence. Furthermore, increased agreement does not necessarily imply improved correctness; it remains possible that reasoning introduces shared biases across annotators. Future work should evaluate alignment with gold labels or expert adjudication to distinguish improved understanding from correlated error.

Future work should extend this framework to larger and more diverse datasets, explore alternative forms of reasoning presentation, and incorporate direct evaluation of reasoning quality and annotator trust. More broadly, we aim to disentangle whether observed improvements arise from better interpretive cues or shared biases introduced by the model.

\section{Acknowledgments}
This work is supported in part by the National Science Foundation award number 2343387.

\bibliography{custom}







\end{document}